# Estimating Parkinsonism Severity in Natural Gait Videos of Older Adults with Dementia

Andrea Sabo, Sina Mehdizadeh, Andrea Iaboni, Babak Taati

*Abstract*— Drug-induced parkinsonism affects many older adults with dementia, often causing gait disturbances. New advances in vision-based human pose-estimation have opened possibilities for frequent and unobtrusive analysis of gait in residential settings. This work leverages spatial-temporal graph convolutional network (ST-GCN) architectures and training procedures to predict clinical scores of parkinsonism in gait from video of individuals with dementia. We propose a two-stage training approach consisting of a self-supervised pretraining stage that encourages the ST-GCN model to learn about gait patterns before predicting clinical scores in the finetuning stage. The proposed ST-GCN models are evaluated on joint trajectories extracted from video and are compared against traditional (ordinal, linear, random forest) regression models and temporal convolutional network baselines. Three 2D human pose-estimation libraries (OpenPose, Detectron, AlphaPose) and the Microsoft Kinect (2D and 3D) are used to extract joint trajectories of 4787 natural walking bouts from 53 older adults with dementia. A subset of 399 walks from 14 participants is annotated with scores of parkinsonism severity on the gait criteria of the Unified Parkinson's Disease Rating Scale (UPDRS) and the Simpson-Angus Scale (SAS). Our results demonstrate that ST-GCN models operating on 3D joint trajectories extracted from the Kinect consistently outperform all other models and feature sets. Prediction of parkinsonism scores in natural walking bouts of unseen participants remains a challenging task, with the best models achieving macro-averaged F1-scores of 0.53 ± 0.03 and 0.40 ± 0.02 for UPDRS-gait and SAS-gait, respectively. Pre-trained model and demo code for this work is available: https://github.com/TaatiTeam/stgcn_parkinsonism_prediction.

## I. INTRODUCTION

PARKINSONISM is a movement disorder primarily characterized by tremor, bradykinesia, and gait disturbances [1]. Drug-induced parkinsonism (DIP) is the most common cause of parkinsonism after Parkinson's disease (PD) [2]. Approximately 30 to 60% of individuals with dementia treated with conventional antipsychotics experience DIP [3], [4]. A similar prevalence of DIP is associated with high doses of atypical antipsychotics in this population [4]. It is important to identify worsening DIP in a timely manner so that appropriate action (such as reducing or discontinuing the causal medication) can be taken to manage the condition.

Gait disturbances associated with DIP are difficult to monitor as they are subject to short-term fluctuations that are not well-captured by formal gait assessments. Formal gait assessments are performed by trained professionals during infrequent clinical visits in which parkinsonism severity in gait is rated on the gait criteria of the Unified Parkinson's Disease Rating Scale (UPDRS) and Simpson-Angus Scale (SAS) [5], [6]. These scales score the severity of impairment on integer scales from 0 to 4, with a higher score indicating more severe impairment. Therefore, there is an opportunity for longitudinal monitoring of parkinsonian symptoms in residential settings to compliment formal clinical assessments.

Previous work has proposed the use of smartwatches to monitor dyskinesia and resting tremor in home settings in a population of adults with PD [7]–[9]. However, the compliance of a wearable smartwatch system has been found to be low in this population over periods longer than 6-13 weeks [9]. Conversely, zero-effort technologies, such as camera-based ambient monitoring systems which do not require any action on the part of the participant (such as charging or putting on a device), are more feasible for collecting longitudinal data in older adults and particularly those with cognitive impairment.

Multi-view video data has also been explored for gait analysis applications. For example, work by Kwolek et al.

†This work was supported by the Walter and Maria Schroeder Institute for Brain Innovation and Recovery; KITE, Toronto Rehabilitation Institute; Canadian Institute of Health Research (CIHR), National Sciences and Engineering Research Council (NSERC, Canada) discovery grant (RGPIN 435653); Alzheimer's Association (USA) & Brain Canada (New Investigator Research Grant NIRG-15-364158); AMS Healthcare (AMS Fellowship in Compassion and Artificial Intelligence); the Vector Scholarship in Artificial Intelligence; and the Ontario Graduate Scholarship.

A. Sabo is with KITE, Toronto Rehabilitation Institute, University Health Network (UHN), Toronto, Canada; and the Institute of Biomedical Engineering, University of Toronto, Toronto, Canada. (e-mail: Andrea.Sabo@mail.utoronto.ca)
S. Mehdizadeh is with KITE, Toronto Rehabilitation Institute, UHN, Toronto, Canada. (e-mail: Sina.Mehdizadeh@uhnresearch.ca)
A. Iaboni is with KITE, Toronto Rehabilitation Institute, UHN, Toronto, Canada; the Department of Psychiatry, University of Toronto, Toronto, Canada; and the Centre for Mental Health, UHN, Toronto, Canada. (e-mail: Andrea.Iaboni@uhn.ca)
B. Taati (corresponding author) is with KITE, Toronto Rehabilitation Institute, UHN, Toronto, Canada; the Department of Computer Science, University of Toronto, Toronto, Canada; and the Institute of Biomedical Engineering, University of Toronto, Toronto, Canada. (email: Babak.Taati@uhn.ca)

presents a dataset of 166 walking sequences recorded using 10 motion capture (moCap) cameras and 4 synchronized video cameras [10]. The authors note that while there were differences in the predicted positions of key joints extracted using moCap and standard color video, they were able to achieve comparable results for the downstream task of gait identification with appropriate predictive models. A similar methodology was used to collect moCap and multi-view video data as part of the Human3.6M dataset, which includes recordings of gait in addition to other activities [11]. Such datasets are valuable in understanding how joint position estimations from video compare to known high accuracy moCap systems.

While multi-view video data collected using commercial 3D motion analysis systems (eg. Optitrack or Vicon) have been successfully used for clinical gait assessment [12], these systems require multiple carefully calibrated cameras and are thus not practical for use in residential settings. In contrast, single-camera systems are well-suited for assessment of parkinsonian gait in residential settings as one camera can be used to monitor the entire body unobtrusively.

Previous work on vision-based gait assessment has explored the use of Microsoft Kinect sensors for analysis of parkinsonian gait using the 3D joint positions provided by the system [13], [14]. Due to the onboard time-of-flight sensor, an advantage of Kinect v2 devices is that they can be used in low-light environments. However, due to the technical limitations of the Kinect depth sensor, 3D joint positions can only be extracted when the participant is between 0.5 – 4.5 m from the sensor, thus limiting the number of gait cycles that can be analyzed with a single sensor [15]. To expand the depth of field and facilitate the recording of longer walking bouts, multi-Kinect systems have been proposed [13], [16]. While suitable for clinical use, multi-Kinect systems cost more and must be installed precisely, and are thus not feasible for residential use.

Conversely, standard color video does not rely on depth sensors, allowing longer walking bouts to be analyzed using a single camera. Recent advances in computer vision and machine learning algorithms have enabled more comprehensive and automated analysis of video recorded on consumer-grade devices. Specifically, human-pose estimation libraries such as OpenPose, Detectron, and AlphaPose have enabled extraction of 2D joint pixel positions from color video [17]–[20]. Related studies have examined the use of 2D joint trajectories for computing domain specific features for identifying parkinsonian gait and rating dyskinesias from color video [21]–[24]. Similarly, work by Lu et al. has explored the use of 3D joint trajectories extracted from video to predict scores of parkinsonism in gait [25]. However, the performance of human pose-estimation libraries is evaluated on benchmarks such as MPII Human Pose or the COCO Keypoint Detection Task [26], [27], so there is limited understanding about whether better performance on these benchmarks leads to better performance on downstream tasks.

Furthermore, spatial temporal graph convolutional networks (ST-GCNs), which leverage the innate graph structure of the human body skeleton, offer an effective mechanism to learn directly from joint trajectories [28]. The advantage of these models is that engineered gait features no longer need to be developed and calculated from the joint trajectories first, as the ST-GCN can learn to use the most important aspects of the gait pattern directly from the joint trajectories. ST-GCNs have been successfully combined with human pose-estimation libraries for human action recognition, as well as for scoring leg agility on the UPDRS [28], [29]. However, the use of these models for evaluating parkinsonism in gait on clinical scales remains unexplored.

This work explores the potential and challenges of using standard color video to score parkinsonian gait in non-clinical settings. We evaluate three ST-GCN architectures to predict UPDRS-gait and SAS-gait scores in natural walks of unseen participants. The ST-GCN models evaluated in this work are based upon previous models proposed by Yan et al. [28] and will be benchmarked against traditional regression models (linear, logistic, random forest) and temporal convolutional networks. Furthermore, 3D joint trajectories extracted from a Microsoft Kinect sensor are compared to 2D joint trajectories extracted from color video using open-source human pose-estimation libraries (OpenPose, Detectron, AlphaPose) to determine if information from the additional spatial dimension improves prediction of clinical scores when used as input for the proposed models. This work is the first to investigate whether there are significant differences in prediction of UPDRS-gait and SAS-gait score when each of these three 2D pose-estimation libraries are used to extract input joint trajectories. Overall, this study presents a comprehensive analysis of how the selection of machine learning model architecture and training methodology, as well as the input data used, affects prediction of clinical scores of parkinsonism in gait in older adults with dementia.

## II. METHODS

### A. Data Collection

The core of the data used in this investigation was collected as part of a larger observational study at the dementia in-patient unit of the Toronto Rehabilitation Institute – University Health Network (UHN). Participants with a diagnosis of dementia and the ability to walk 20 m independently (without a walking aid) were recruited for the study. The Research Ethics Board at UHN approved the protocol (protocol number 15-9693) and substitute decision makers provided consent for study enrolment. Participants provided assent for data collection.

Natural walking bouts were recorded using AMBIENT, a ceiling-mount Microsoft Kinect v2 system (1080 × 1920 pixels, 30 Hz) in a hallway of the hospital. To preserve the privacy of other individuals on the unit, a radio-frequency identification (RFID) system was used to only begin recording video when study participants entered the hallway. A detailed description of the AMBIENT system is available in [23]. The AMBIENT system recorded natural walking bouts of participants over multiple weeks during their stay at the unit. The captured videos were sorted to exclude walking bouts in which a participant used the handrail located along the wall of the hallway or was not walking independently.

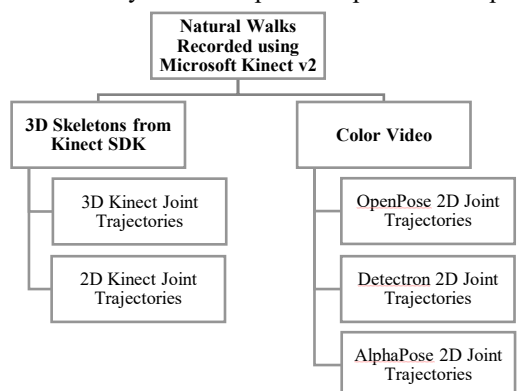

Fig. 1. Joint trajectory feature sets extracted from recorded walks.

To supplement the core AMBIENT dataset and increase the size of the training dataset, videos of additional walking bouts were recorded at an independent living facility for older adults. The participants recruited for this study did not have a diagnosis of dementia or any physical or neurological conditions that affected gait. Participants (healthy older adults aged 65 years and above) were instructed to continuously walk towards and away from stationary mobile phone cameras (1080 × 1920 pixels, 30 Hz) for one minute. The Research Ethics Board at the University of Toronto approved the protocol (protocol number 00037569) for this study and the participants provided written consent prior to data collection.

Walks from 14 of the 53 participants recorded using the AMBIENT system were scored on the UPDRS-gait and SAS-gait scales by a specialist. Due to the large number of recorded videos, it was not possible to annotate all walks so participants were selected to include a range of parkinsonian severities in the dataset. The walking bouts were scored from the Kinect video recordings. When multiple walks of a participant were captured on the same day, one bout was randomly selected for annotation to avoid biasing the dataset with many similar walks recorded in close succession.

### B. Extraction of Joint Trajectories and Calculation of Gait Features

The Kinect sensor used as part of the AMBIENT system simultaneously recorded standard color videos and captured depth information, allowing 3D skeleton trajectories to be extracted using the Kinect software development kit (SDK). As summarized in Fig. 1, the standard color videos and 3D skeleton data were processed independently to extract 2D and 3D joint trajectories and gait features for each walk.

*1) Extraction of 2D Joint Trajectories and Gait Features from Color Video*

All standard color videos (from both the dementia in-patient unit and the independent living facility) were processed with three human pose-estimation libraries (OpenPose, Detectron, AlphaPose). These libraries were used to obtain estimates of x and y joint pixel positions as well as confidence scores of the body joints in each frame of the video [17]–[20]. Table 1 lists the model backbones and output of each library. Because the videos were recorded with the participants walking towards a ceiling or tripod-mounted camera, the extracted 2D coordinates roughly correspond to the frontal plane of motion. As also described by Stenum et al. [30], there were instances in which a joint on the left side of the body was mistaken for the analogous joint on the right side of the body by the pose-estimation libraries. To avoid propagating the errors introduced by these left/right mislabels to subsequent interpolation and filtering steps, the errors were manually corrected by switching the left/right labels. The resultant joint trajectories were then filtered with a zero phase, second order low-pass Butterworth filter with a cut-off frequency of 8 Hz.

Footfall timings were automatically identified by differentiating the vertical positions of the ankle keypoints and denoting the heel strikes at 35% of the peak value of each gait cycle, as proposed in [31]. After identifying the heel strikes, seven spatiotemporal and mechanical stability features of gait (cadence, number of steps, average step width, average margin of stability, the coefficient of variation of step width and time, and the symmetry index of step times) were calculated. A complete description of the methods for detecting heel strikes and calculating these 2D gait features is presented by Ng et al [32]. Finally, the section of each walk between the first and last step was extracted using the detected footfall times, ensuring that the participant was continuously ambulating during the final joint trajectory. This joint trajectory and gait feature extraction process was repeated independently for each pose-estimation library, yielding three feature sets, one for each of OpenPose, Detectron, and AlphaPose.

*2) Processing of 3D Joint Trajectories from Microsoft Kinect*

The Kinect SDK also provides 3D joint trajectories when the participant is within the operating range of the depth sensor. These 3D joint trajectories were similarly filtered with a zero phase, second order, low-pass Butterworth filter with a cut-off frequency of 8 Hz. A total of 16 spatiotemporal, variability, symmetry, and mechanical stability features of gait were calculated directly from the 3D joint trajectories. These gait features are described and validated in [33]–[37]. Note that unlike the 2D joint trajectories extracted using the pose-estimation libraries, the 3D joint positions extracted from the Kinect are provided in meters, thus allowing distance measures to be associated with real-world units.

Furthermore, to create an additional 2D dataset, the 3D Kinect joint trajectories were projected back into 2D camera space using the intrinsic matrices of the depth and color cameras. Using the same method as for the 2D joint trajectories obtained using pose-estimation libraries, the seven previously described 2D features of gait were also calculated for this dataset.

*3) Formatting of Joint Trajectories for Input to Machine Learning Models*

To account for the varying locations of the participants within the field of view of the camera, the hip center of all joint trajectories was set at (x, y, [z – for 3D only]) values of (100, 100, 100) in each frame. Thirteen keypoints (nose, shoulders, elbows, wrists, hips, knees, and ankles) were selected from the joint trajectories to ensure consistency between the data obtained from all pose-estimation libraries as well as from the Kinect SDK. Furthermore, all joint trajectories were mirrored along the vertical axis to double the size of the dataset and provide more training data for the machine learning models.

TABLE 1
MODEL DETAILS, OUTPUT, AND CONFIDENCE SCORE THRESHOLD FOR INTERPOLATION FOR EACH HUMAN POSE-ESTIMATION LIBRARY

| Pose-Estimation Library | Model Details | Output |
|---|---|---|
| OpenPose | Release v1.5.1 | 25 joint keypoints |
| Detectron | ResNet101-FPN backbone | 17 joint keypoints |
| AlphaPose | YOLOv3-spp detector, pretrained ResNet-50 backbone | 17 joint keypoints |

TABLE 2
FILTER COUNT FOR TEMPORAL CONVOLUTIONAL MODELS

| Model | Filter Count per Layer |
| --- | --- |
| TCN_1 | [64*, 128, 128, 256, 256**] |
| TCN_2 | [64*, 128, 128, 256, 256, 256**] |
| TCN_3 | [32*, 64, 64, 128, 128**] |
| TCN_4 | [128*, 128, 256, 512, 512**] |

*3D conv layer; **Fully connected linear layer; all other layers are 1D conv

TABLE 3
FILTER COUNT FOR ST-GCN MODEL BACKBONES

| Model | Filter Count per Layer |
| --- | --- |
| ST-GCN_Small | [16, 16, 32, 32] |
| ST-GCN_Medium | [32, 32, 64, 64] |
| ST-GCN_Large | [64, 64, 64, 64, 128, 128, 128, 256, 256, 256] |

The calculated gait features and clinical scores were the same for the mirrored walks.

### C. Prediction of Clinical Scores

Machine learning models were trained to predict UPDRS-gait and SAS-gait scores from the extracted gait features and joint trajectories. Baselines consisting of traditional regression models and temporal convolutional networks were first established to allow for direct comparison to the proposed spatiotemporal graph convolutional network models on our dataset.

To simulate testing on unseen participants, all models were evaluated using a leave-one-subject-out cross-validation (LOSOCV) scheme whereby all walks from the participant being evaluated were excluded from the training and validation sets. The primary metric used to quantify model performance was the macro-averaged F1 score to avoid bias due to class imbalance in the dataset.

#### 1) Baseline Models

To establish a baseline, traditional regression models operating on gait features were trained to predict UPDRS-gait and SAS-gait scores for the annotated walking bouts on the subset of 14 participants. The four models investigated were linear and random forest regression models (Scikit-Learn toolbox [38]) as well as the immediate and absolute threshold versions of ordinal logistic regression (Mord Python package [39]). Hyperparameter tuning was performed using 10-fold cross validation with 1000 random search iterations. Macro-averaged F1 score on the validation set was used to select the best model for evaluation on the held-out participant.

Additionally, temporal convolutional networks (TCNs) trained to predict UPDRS-gait and SAS-gait from joint trajectories were investigated as another baseline model. The input to the TCN baseline models was a $\mathbb{R}^{N \times D \times T}$ tensor where N is the number of joints (13 in this study), D is the number of dimensions (two for 2D data, three for 3D data), and T is the number of timesteps. The number of timesteps was selected to be 120 (representing 4 seconds of video at 30 Hz), ensuring that most walks in the dataset had at least this many temporal data points. Joint trajectories shorter than 120 timesteps were zero-padded, while longer trajectories in the training set were sampled to select a different 120 timestep sequence for training in each epoch. For trajectories in the validation and test sets, the center 120 timesteps in each sequence were used to avoid introducing an additional source of variability in the evaluation metrics between epochs.

The first layer of the baseline TCN models was a 3D convolutional layer that combines information across all joints for each timestep of the input sequence. Thus, the input to this layer was a tensor of shape $\mathbb{R}^{N \times D \times T}$, and the output was a vector of shape $\mathbb{R}^{1 \times T}$ for each filter. Next, 1D convolutional layers were used to combine information across timesteps. The size of these 1D filters represents the temporal kernel size, with temporal kernel sizes of 5, 9, and 13 investigated in the experiments. Next, the outputs of the last 1D convolutional layer were flattened and used as input to a fully connected layer. The fully connected layer with one output neuron was used to regress to the clinical score of interest. Rectified linear activation (ReLU) was used between all layers, and dropout was used between all 1D convolutional and linear layers. Six dropout rates were investigated, ranging from 0 to 0.5 in increments of 0.1. Four different TCN model architectures were investigated and are presented in Table 2.

Finally, the Ordinal Focal Double-Features Double-Motion Network (OF-DDNet) model proposed by Lu et al. [25] was used to predict parkinsonism scores on this dataset. A key characteristic of this model is that it formulates the task as an ordinal classification problem rather than a regression problem. This model was evaluated using the method described in the original work (using VIBE [40], a 3D pose-estimation library operating on color video) and using 3D Kinect joint trajectories as input.

#### 2) Spatiotemporal Graph Convolutional Networks (ST-GCNs)

As an alternative to the traditional regression and TCN baseline models, spatiotemporal graph convolutional networks (ST-GCNs) that operate on joint trajectories were investigated for prediction of clinical parkinsonism scores. ST-GCN models operate on trajectories of joint positions, however, rather than using a 3D filter that combines information across all joints at each time step as is done in TCNs, the ST-GCN models use filters that only consider neighbouring joints in the skeleton in the spatial dimension, allowing ST-GCNs to better leverage domain knowledge about which joints are connected to each other [21]. Multiple layers of ST-GCN blocks are used to increase the spatial and temporal receptive fields of the models to cover the entire input joint trajectory. Using the same method as for the TCN models, a 120 timestep trajectory of 13 joints was used as input to the ST-GCN models explored in this study.

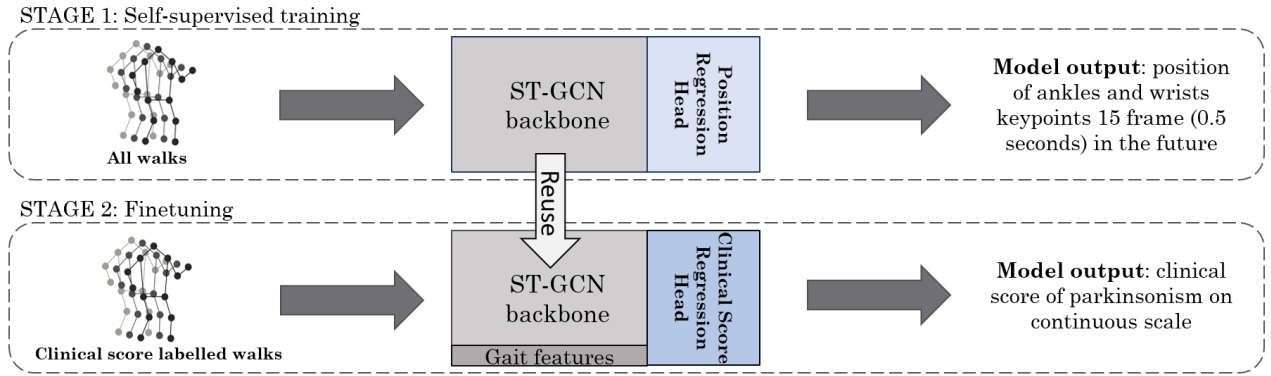

Fig. 2. Two-stage training method for ST-GCN models. The use of gait features is optional.

Three ST-GCN backbone architectures were evaluated: the original 10-layer model developed by Yan et al. [28] for human action recognition, as well as two smaller, 4-layer models developed with the goal of reducing overfitting due to the smaller dataset size available in this study. The 4-layer models investigated in this work were selected through preliminary empirical evaluation of 10 candidate models. Table 3 presents the filter counts for each of the selected models.

*a)    ST-GCN Self-Supervised Training*

As summarized in Fig. 2, the ST-GCN models were trained in two stages; a self-supervised pretraining stage, followed by a supervised fine-tuning stage. The aim of the self-supervised pretraining stage was to first encourage the ST-GCN model to learn the underlying patterns of gait before being used to predict the clinical scores. Because this pretraining task did not require clinical score labels, all walks from all 53 participants were used in this stage. For the pretraining task, a fully connected layer with $J \times D$ outputs was added on top of the ST-GCN backbone. This position regression head was used to predict the position of $J$ joints across the $D$ dimensions of the input data. In the experiments, four joints (left/right ankles and wrists) were predicted 15 frames (0.5 s) after the end of the input sequence. During training, the Wing loss between the true and predicted locations of the joints was minimized [41].

*b)    ST-GCN Finetuning to Predict Clinical Scores*

To predict clinical scores of parkinsonism, the position regression head was replaced with a clinical score regression head consisting of a fully connected layer with a single output. The ST-GCN backbone weights were initialized with those from the output of the pretraining stage. The weights were not frozen and could be updated during finetuning. Preliminary experiments in which the ST-GCN backbone weights were frozen yielded significantly poorer results and were thus not investigated further. A variation of the ST-GCN models in which the pre-computed gait features were combined with the ST-GCN features immediately before the fully connected layer was used to investigate the effect of including gait features in addition to joint trajectories.

During finetuning, a two-part loss was used. For walks with true clinical score labels, the score loss ($\mathcal{L}_{score}$) was calculated as the mean squared error (MSE) between the true and predicted clinical score. To address class imbalance in the clinical score labels, the score loss was weighted by the inverse of the class prevalence, thus giving equal weighting to all classes. Additionally, a flip loss ($\mathcal{L}_{flip}$) was calculated as the MSE between the predicted score of a joint trajectory and the predicted score assigned to the vertical mirror of the same joint trajectory. Because the flip loss does not require annotations of clinical scores, it can be calculated for all walks. These two individual loss terms were combined to form an overall loss:

$$\mathcal{L}_{total} = \mathcal{L}_{score} + \alpha \mathcal{L}_{flip} \qquad (1)$$

*c)    ST-GCN Experimental Design and Ablations*

A large initial experiment was first performed to assess the impact of model backbone, kernel size, dropout, and feature set on prediction of clinical scores. Three temporal kernel sizes (5, 9, 13) and six dropout rates (ranging from 0 to 0.5 in increments of 0.1) were investigated for each of the three ST-GCN backbones presented in Table 3. Models were trained on joint trajectories with and without gait features. Overall, 108 hyperparameter configurations were tested for each clinical score and each of the 5 feature sets (OpenPose, Detectron, AlphaPose, 2D Kinect, 3D Kinect). Self-training to predict future joint positions was performed and an α of 0 was used in the finetuning loss. This selection of α was used to omit the flip loss, significantly reducing the training time for the models as only walks with clinical labels were required for training. The examination of the flip loss was reserved for the ablation experiments. Based on these experiments, the top three models from each feature set (with and without gait features) with the highest macro-averaged F1-score across the validation set were selected for the ablation experiments.

An ablation experiment exploring the use of the flip loss during finetuning was performed. Due to the large number of computational resources required to train the models on all walks (rather than only those with clinical labels), a flip loss with α=1 was evaluated.

Two ablation experiments were performed to evaluate the impact of pretraining the ST-GCN models using the self-supervised task. In the first experiment, the effectiveness of the

TABLE 4
DEMOGRAPHIC DATA OF STUDY PARTICIPANTS AND NUMBER OF VIDEOS OF WALKING BOUTS PER MODALITY AND DATA COLLECTION LOCATION

|  | AMBIENT (all) | | AMBIENT (clinical labels) | | Independent Living Facility |
|---|---|---|---|---|---|
|  | *3D depth data* | *Color videos* | *3D depth data* | *Color videos* | *Color videos* |
| Number of participants | 53 | 49 | 14 | 14 | 14 |
| Age (years ± SD) | 76.5 ± 7.9 | 76.5 ± 8.0 | 76.2 ± 8.7 | 76.2 ± 8.7 | 86.7 ± 6.2 |
| Sex (% male) | 56.6 | 57.1 | 57.1 | 57.1 | 21.4 |
| Height (cm ± SD) | 163.8 ± 14.3 | 163.8 ± 14.8 | 159.2 ± 22.8 | 159.2 ± 22.8 | 165.6 ± 10.0 |
| Weight (kg ± SD) | 66.6 ± 12.9 | 66.2 ± 13.0 | 62.7 ± 14.0 | 62.7 ± 14.0 | 64.1 ± 12.5 |
| Number of recorded walking bouts per participant ± SD (total) | 90.3 ± 71.3 (4787) | 54.6 ± 65.0 (2741) | 28.6 ± 12.1 (400) | 25.9 ± 14.4 (362) | 9.6 ± 2.1 (134) |

*SD = standard deviation

TABLE 5
NUMBER OF EXTRACTED WALKING BOUTS PER FEATURE SET

| Feature Set | AMBIENT | | | Independent Living Facility | Total |
|---|---|---|---|---|---|
| | *Annotated walks with skeleton trajectories* | *Annotated walks with gait features* | *All walks with skeleton trajectories* | *All walks with skeleton trajectories (no annotations)* | *All walking bouts* |
| OpenPose | 312 | 310 | 2025 | 132 | 2157 |
| Detectron | 330 | 313 | 2115 | 134 | 2249 |
| AlphaPose | 324 | 321 | 2160 | 132 | 2292 |
| Kinect 2D | 399 | 76 | 4778 | 0 | 4778 |
| Kinect 3D | 399 | 399 | 4778 | 0 | 4778 |

pretraining task in learning to predict the positions of the wrists and ankles 0.5 s in the future was validated by comparing the mean absolute error (MAE) of the positions predicted by the models and the last positions of the joints in the input sequence. To confirm whether the use of the pretraining task improves prediction of clinical scores in the downstream task, an additional experiment was performed in which the pretraining stage was omitted. In this experiment, the weights for the finetuning stage set using Xavier initialization instead.

Finally, an ablation experiment in which data from all three 2D pose-estimation libraries (AlphaPose, Detectron, OpenPose) were used to train a single ST-GCN. The goal of this experiment was to use data from other pose-estimation libraries as augmentations for each walking bout and evaluate the impact on model performance.

*3) Convolutional Model Training Details*

All convolutional models (TCN and ST-GCN) were trained using PyTorch using a cyclic learning rate. The learning rate range was annealed by a factor of ten every 20 epochs. Training was terminated when the loss on the validation set did not decrease for 25 consecutive epochs. A training curriculum in which the models were only exposed to walks at the extrema of the available scores for each criterion was used for the first 25 epochs of the finetuning stage. This was done to encourage the model to first learn about examples of walks at the ranges of the scales. After the 25[th] epoch, the models were trained on all available walks. Because the final output layer predicted the clinical score on a continuous scale, the final prediction was rounded the nearest integer and clipped to be between 0 and highest clinical score label in the training set. A batch size of 100 was used. As with the traditional regression models, the convolutional models were trained using LOSOCV. Five-fold cross-validation was used, dividing the walks not from the held-out participant into a 80/20 train/validation split. Experimental results were logged using Weights and Biases [42].

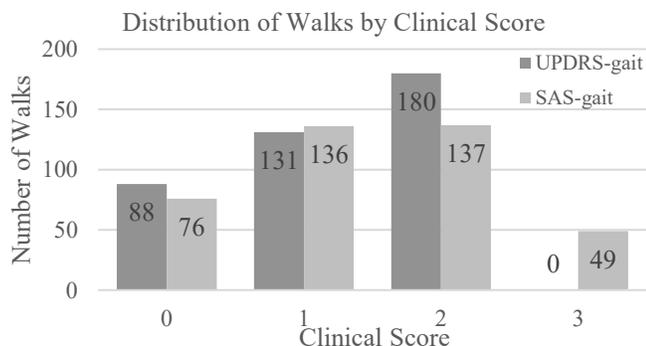

Fig. 3. Number of walks per clinical score.

Data augmentation in which joint trajectories were stretched and skewed to simulate different camera angles and body shapes were initially explored, but hindered model performance and were thus omitted. Similarly, normalization of joint trajectories by hip or shoulder pixel distance in each frame resulted in worse model performance and was also excluded.

### III. RESULTS AND DISCUSSION

Videos of 53 participants were collected using the AMBIENT system. Data from an additional 14 participants were collected at the independent living facility. Table 4 summarizes the demographic information about the entire AMBIENT dataset, the subset that was annotated with clinical scores, and the data collected at the independent living facility. Data collected as part of the AMBIENT dataset was recorded using one of two modes: a depth mode which only allowed extraction of 3D skeleton trajectories and a color video + depth mode that also allowed extraction of joint trajectories from color video. Therefore, 3D skeleton trajectories were available for all walks, while color video was only available for a subset.

Table 5 presents the number of skeleton trajectories extracted from the recorded walking bouts. Gait features were only calculated for annotated walks in which more than 3 steps were detected. The distribution of UPDRS-gait and SAS-gait scores of the annotated walks from the subset of 14 participants is presented in Fig. 3.

### A. Hyperparameter Search and Comparison of Baseline and Proposed ST-GCN Models

As previously described, a large hyperparameter search for the baseline (traditional and TCN) and ST-GCN models was conducted. The macro F1-score on the test set for the traditional baseline regression models operating on gait features, the baseline TCN models operating on joint trajectories, and ST-GCN models operating on joint trajectories (and optionally gait features) are presented in Table 6. For the TCN and ST-GCN models, Table 6 presents the mean and standard deviation of the test macro F1-score for each of the 5 folds. Note that all models were selected according to the highest macro F1-score on the validation set; so there are configurations for which the macro F1-score was higher on the test set than the values presented in the table.

As presented in Table 5, the number of walks from the Kinect 2D feature set for which gait features were extracted was significantly lower than the other 2D pose-estimation libraries. The Kinect is only able to extract skeleton data when the participant is between 0.5 – 4.5 m from the sensor. When the Kinect 3D data is projected to camera space and processed with the 2D gait feature extraction pipeline, most walks have less than 3 detected steps and gait features thus cannot be calculated. The is not an issue with Kinect 3D feature set because the gait feature extraction pipeline leverages depth information, and is thus able to consistently detect at least 3 steps in the same data. For this reason, metrics for traditional regression models that rely solely on gait features are not presented for the Kinect 2D feature set.

TABLE 6
MACRO F1-SCORE OF BEST BASELINE AND ST-GCN MODELS ON TEST DATA

| Feature Set | Test macro F1-score of top models – UPDRS-gait | | | | Test macro F1-score of top models – SAS-gait | | | |
|---|---|---|---|---|---|---|---|---|
| | *Traditional regression models (gait features)* | *TCN (no gait features)* | *ST-GCN (no gait features)* | *ST-GCN (with gait features)* | *Traditional regression models (gait features)* | *TCN (no gait features)* | *ST-GCN (no gait features)* | *ST-GCN (with gait features)* |
| OpenPose | 0.293 | 0.193±0.082 | 0.282±0.032 | 0.322±0.027 | 0.232 | 0.199±0.052 | 0.236±0.026 | 0.250±0.028 |
| Detectron | 0.344 | 0.157±0.005 | 0.384±0.034 | 0.369±0.030 | 0.236 | 0.183±0.032 | 0.292±0.044 | 0.303±0.029 |
| AlphaPose | 0.333 | 0.160±0.003 | 0.321±0.013 | 0.372±0.019 | 0.230 | 0.224±0.035 | 0.283±0.059 | 0.301±0.014 |
| Kinect 2D | N/A | 0.227±0.02 | 0.327±0.039 | 0.328±0.016 | N/A | 0.241±0.016 | 0.319±0.031 | 0.330±0.031 |
| Kinect 3D | 0.387 | 0.231±0.056 | 0.380±0.024 | 0.492±0.015 | 0.235 | 0.284±0.057 | 0.341±0.015 | 0.332±0.034 |

TABLE 7
MACRO F1-SCORE OF OF-DDNET MODEL [25] TRAINED ON VIBE AND KINECT 3D JOINT TRAJECTORIES

| Feature Set | Test macro F1-score of top models – UPDRS-gait | Test macro F1-score of top models – SAS-gait |
|---|---|---|
| **OF-DDNet [25]** | | |
| VIBE –pretrained model [25] | 0.296 | N/A |
| VIBE – trained on our data | 0.364 ± 0.075 | 0.296 ± 0.034 |
| Kinect 3D – trained on our data | 0.416 ± 0.032 | 0.234 ± 0.034 |
| **Ours – Best ST-GCN** | | |
| Kinect 3D – without self-supervised pre-training | 0.492 ± 0.015 | 0.341 ± 0.015 |
| Kinect 3D – with self-supervised pre-training | 0.526 ± 0.026 | 0.398 ± 0.018 |

As presented in Table 6, traditional regression models operating only on engineered gait features outperform TCNs operating on a 120 timestep sequence of joint trajectories. However, when the joint trajectories were used as input to ST-GCNs, the macro-averaged F1 scores on the test set improved across all feature sets for all clinical scores when compared to the traditional or TCN models. These findings confirm that ST-GCN are able to better leverage the spatial structure of skeleton trajectories than TCNs and are thus better suited for prediction of clinical scores of parkinsonism in gait.

Furthermore, our results suggest that the benefit of including gait features in addition to joint trajectories as input to ST-GCN models is dependent on the clinical score being predicted, as well as on the feature set. Gait features improve prediction of UPDRS-gait using the OpenPose, AlphaPose, and Kinect 3D feature sets, but hinder or do not significantly change the prediction of UPDRS-gait scores using the Detectron and Kinect 2D feature sets (Table 6). Conversely, when predicting SAS-gait scores, including gait features (in addition to joint trajectories) does not significantly affect prediction performance of the ST-GCN models across all feature sets.

Comparing the performance of the ST-GCN models in Table 6 across different feature sets, the top performing models for prediction of both clinical scores operated on the Kinect 3D feature set. Of the five feature sets investigated in this study, the Kinect 3D feature set was the only one that captured information from the depth dimension in both the joint trajectories and gait features. The inclusion of the depth dimension also allows for the calculation of additional gait features that rely on data in the anterior-posterior direction (such as step length and gait speed) for the Kinect 3D feature set. For prediction of UPDRS-gait scores, the larger set of gait features part of the Kinect 3D feature set resulted in a test macro F1-score of 0.387, higher than all models tested on the four 2D feature sets. However, the same trend was not observed in for the prediction of SAS-gait scores, suggesting that gait features capture less information about the SAS-gait score.

Amongst the 2D feature sets, the top models used the data extracted using human pose-estimation libraries operating on color video, as opposed to the Kinect 2D feature set. This suggests that with respect to 2D data, the currently available human pose-estimation libraries operating on standard color video outperform 2D projections of 3D data obtained from specialized depth cameras such as the Microsoft Kinect. Therefore, the superior results obtained with the Kinect 3D feature set can be attributed to the additional depth information it captures, rather than better 2D pose-tracking performance in the camera space.

Comparing the results from the ST-GCN models (Table 6) to the OF-DDNet model (Table 7) proposed by Lu et al. [25], the highest macro-averaged F1-score was reported by an ST-GCN model for both clinical scores. Interestingly, the joint trajectories extracted from video using VIBE yielded a higher macro F1-score than the Kinect 3D feature set for prediction of SAS-gait (Table 7), but when an ST-GCN model was used to with the same Kinect 3D data, the F1-macro average score was higher.

### B. ST-GCN Ablation Experiments

After confirming that the proposed ST-GCN models predict clinical scores of parkinsonism better than baseline regression and TCN models, ablation experiments were performed to evaluate the effect of the flip loss in the finetuning stage, training on data from multiple pose-estimation libraries, and the self-supervised pretraining stage.

#### 1) Flip Loss Ablation

The top three models for each feature set and clinical score were also used to assess the effect of the flip loss ($\mathcal{L}_{flip}$) during finetuning. Using α=1 for the Stage 2 loss in Eqn. 1, the macro-averaged F1-scores on the test sets for the best models (as determined using validation macro F1-scores) are presented in Table 8. The results from this ablation experiment suggest that there is a benefit to training with a flip loss (Table 8) as opposed to excluding it (Table 6). For all feature sets excluding OpenPose, the top performing model for each clinical score was trained with a α=1 flip loss term (Eq. 1). The purpose of this flip loss is to encourage the models to predict the same score for a walk and the same walk but mirrored vertically. As the flip loss does not rely on labels of UPDRS-gait or SAS-gait scores, both labelled and unlabelled walks were used to train the models. Because the larger input datasets also increase the time and computational resources required to train the models, it is still beneficial to narrow down the model hyperparameters to evaluate by first performing preliminary experiments that omit the flip loss (and are thus less computationally intensive).

TABLE 8
MACRO F1-SCORE OF BEST ST-GCN MODELS USING SELF-SUPERVISED PRETRAINING AND FLIP LOSS IN FINETUNING STAGE

| Feature Set | Test macro F1-score of top models – UPDRS-gait | | Test macro F1-score of top models – SAS-gait | |
|---|---|---|---|---|
| | *ST-GCN (no gait features)* | *ST-GCN (with gait features)* | *ST-GCN (no gait features)* | *ST-GCN (with gait features)* |
| OpenPose | 0.243±0.019 | 0.265±0.039 | 0.211±0.025 | 0.231±0.028 |
| Detectron | 0.412±0.019 | 0.35±0.015 | 0.317±0.029 | 0.336±0.019 |
| AlphaPose | 0.417±0.028 | 0.385±0.02 | 0.322±0.016 | 0.317±0.019 |
| Kinect 2D | 0.364±0.028 | 0.389±0.025 | 0.311±0.026 | 0.312±0.024 |
| Kinect 3D | 0.395±0.028 | ***0.526±0.026**** | ***0.398±0.018**** | 0.322±0.032 |

\* Top model for clinical score across all experiments

TABLE 9
MACRO F1-SCORE OF ST-GCN MODEL TRAINED ON DATA FROM 2D JOINT TRAJECTORIES EXTRACTED FROM VIDEO (OPENPOSE, DETECTRON, ALPHAPOSE) AND EVALUATED ON ONE POSE-ESTIMATION LIBRARY

| Evaluation Feature Set | Test macro F1-score of top models – UPDRS-gait | | | | Test macro F1-score of top models – SAS-gait | | | |
|---|---|---|---|---|---|---|---|---|
| | *Labelled data only (α=0, no gait features)* | *Labelled data only (α=0, with gait features)* | *All data (α=1, no gait features* | *All data (α=1, with gait features* | *Labelled data only (α=0, no gait features)* | *Labelled data only (α=0, with gait features)* | *All data (α=1, no gait features* | *All data (α=1, with gait features* |
| OpenPose | 0.390 ± 0.015 | 0.368 ± 0.014 | 0.391 ± 0.040 | 0.324 ± 0.051 | 0.269 ± 0.033 | 0.222 ± 0.007 | 0.254± 0.030 | 0.226 ± 0.026 |
| Detectron | 0.441 ± 0.020 | 0.397 ± 0.015 | 0.438 ± 0.049 | 0.358 ± 0.056 | 0.274 ± 0.030 | 0.223 ± 0.012 | 0.269 ± 0.036 | 0.224 ± 0.019 |
| AlphaPose | 0.433 ± 0.028 | 0.400 ± 0.019 | 0.434 ± 0.048 | 0.366 ± 0.042 | 0.305 ± 0.027 | 0.227 ± 0.012 | 0.291 ± 0.033 | 0.225 ± 0.020 |

*2) Simultaneous Training on Data from Three 2D Pose-Estimation Libraries*

The best three ST-GCN models identified in the initial experiments operating on a single feature set were further evaluated by training on data from all three 2D pose-estimation libraries. Each model was trained on data from all three of the OpenPose, Detectron, and AlphaPose feature sets, and then evaluated on data from each feature set individually. LOSOCV was used in the same manner as the previous experiments, ensuring that walks from the test participant did not appear in training or validation for any of the feature sets. The results of this experiment are presented in Table 9.

Comparing the results from Table 9 to those in Table 6, a significant increase in macro F1-score is observed when the model is trained on data from all feature sets for the prediction of UPDRS-gait scores. This trend is not observed for the prediction of SAS-gait scores, where training and predicting on data from only one feature set yielded higher macro F1-scores (Table 6).

These results suggest that the larger dataset size and addition of data augmentation by using different pose-estimation libraries to process each walk is crucial when predicting UPDRS-gait scores with 2D pose-estimation libraries. Interestingly, adding gait features or the flip loss ($\mathcal{L}_{flip}$) does not improve the macro F1-scores as is observed when only one feature set is used for training.

*3) Pretraining Stage Ablation*

Furthermore, the effect of the self-supervised pretraining stage was evaluated on the top three models for each feature set. The effectiveness of the pretraining task for learning about gait patterns was evaluated by evaluating how accurately the pretrained model learned to predict future positions of the wrists and ankles. Table 10 summarizes performance on the pretraining task and presents the MAE on the test set between the true and predicted ankle and wrist positions 15 frames (0.5 s) in the future. The MAE between the true position and the position at the end of the input sequence (15 frames before the predicted timestep) is also presented as a baseline.

To further validate that self-supervised pretraining improves final regression to clinical scores, the pretraining stage was replaced with Xavier initialization of the ST-GCN backbone as in [28] and finetuned with α=1 in the finetuning loss. Using Xavier initialization, the top ST-GCN model achieved a macro-averaged test F1-score of 0.509±0.030 (Kinect 3D, with gait features) for regression to UPDRS-gait scores and 0.384±0.015 (Kinect 3D, no gait features) for regression to SAS-gait scores. These are lower than results obtained with self-supervised pretraining, 0.526±0.026 and 0.398±0.018 respectively, as shown in Table 8.

This experiment also highlighted differences in the underlying skeleton data for the three human pose-estimation libraries operating on video (i.e. OpenPose, Detectron, and AlphaPose). Specifically, as seen in Table 10, the values for the Detectron and AlphaPose rows are very similar for the last position hold columns, suggesting minor differences between the sequences extracted by these two detectors. Conversely, the last position hold values for OpenPose were significantly higher for both ankles and wrists. These findings indicate that for our dataset and preprocessing methodology, the skeleton trajectories in the OpenPose feature set are significantly different than those in the Detectron and AlphaPose feature sets. As seen in Tables 6 and 8, when used as input to the ST-GCN models, the OpenPose feature set results in significantly lower macro-averaged F1-score than the Detectron or AlphaPose feature sets.

TABLE 10
MEAN ABSOLUTE ERROR (MAE) BETWEEN TRUE AND PREDICTED ANKLE AND WRIST KEYPOINTS 15 FRAMES IN FUTURE

| Feature Set | Average MAE on test set | | | | | |
|---|---|---|---|---|---|---|
| | *Stage 1 model prediction – ankles and wrists* | *Last position hold – ankles and wrists* | *Stage 1 model prediction – ankles* | *Last position hold – ankles* | *Stage 1 model prediction – wrists* | *Last position hold – wrists* |
| OpenPose* | 99.7±20.4 | 115.8 | 77.4±16.2 | 97.0 | 121.9±25.8 | 134.6 |
| Detectron* | 30.0±7.1 | 71.2 | 32.5±6.6 | 53.5 | 27.5±7.7 | 88.9 |
| AlphaPose* | 33.7±4.3 | 71.5 | 38.0±4.4 | 54.5 | 29.5±4.6 | 88.4 |
| Kinect 2D* | 60.4±14 | 124.9 | 71.7±14.8 | 115.6 | 49.0±13.4 | 134.1 |
| Kinect 3D** | 16.8±4.1 | 22.7 | 7.5±1.6 | 12.5 | 37.8±10.5 | 35.2 |

\* MAE units: pixels; \*\* MAE units: centimeters

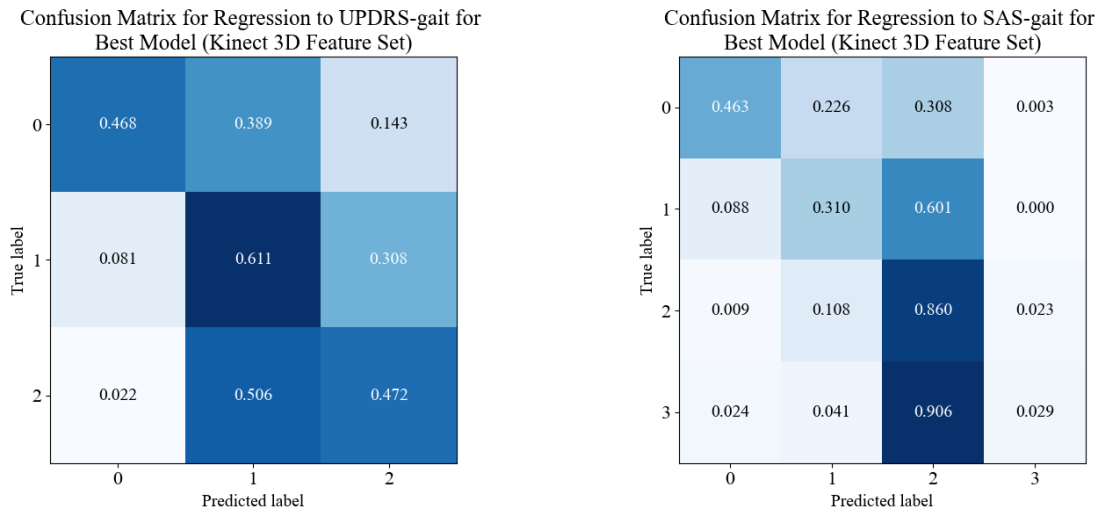

Fig. 4. Confusion matrices for regression to UPDRS-gait (left) and SAS-gait (right) on unseen participants using top models on Kinect 3D feature set. All values are normalized by total true examples of each class.

### C. Summary of Best Models and Confusion Matrices

Overall, the top performing models from our experiments were trained using the Kinect 3D feature set and with the flip loss. The top UPDRS-gait model used the ST-GCN_Medium backbone with a 13 timestep temporal kernel, a dropout of 0.1, and was trained on the Kinect 3D feature set of joint trajectories and gait features. The top SAS-gait model used the ST-GCN_Medium backbone with a 13 timestep temporal kernel, a dropout of 0.0, and was trained on the Kinect 3D feature set of only joint trajectories. Both models were trained with $\alpha=1$ in the Stage 2 loss. Confusion matrices of the top models for each clinical score across all experiments are presented in Fig. 4.

As shown in Fig. 4, the confusion matrix for the top UPDRS-gait model shows that the general trend of clinical scores is relatively well predicted, with misclassifications more likely to occur between adjacent scores on the scale. For the top SAS-gait model, a less prominent pattern is observed. Notably, the model rarely predicts clinical scores of 3. A similar trend was noted for the train set as well, suggesting that the model has difficulty identifying walks at the upper range of the scale. A potential reason for the low model prediction of score 3 could be the lower prevalence of this class in the labelled data (Fig. 3).

## IV. CONCLUSION AND FUTURE WORK

In this work, we evaluated traditional and convolutional models for predicting the clinical scores of parkinsonism in gait using gait features and skeleton trajectories. Our experiments indicate that ST-GCNs trained on skeleton trajectories outperform traditional regression models trained on gait features and TCNs trained on skeleton trajectories. In our experiments, the top models were trained using the Kinect 3D feature set with self-supervised pretraining and a flip loss term in the finetuning stage. With respect to 2D feature sets, models trained using joint trajectories and gait features obtained from the OpenPose human pose-estimation library underperformed models trained on the feature sets extracted using the Detectron and AlphaPose libraries. The best models trained on the Detectron and AlphaPose feature sets outperformed the top models trained on 3D Kinect data that had been projected back to camera space (Kinect 2D feature set), suggesting that specialized depth cameras are not necessary for extracting suitable joint positions in camera space.

Our results suggest that there is an advantage to using feature sets consisting of 3D joint trajectories, so future work will explore the use of human pose-estimation libraries that can predict 3D joint locations using only standard color video. This will eliminate the need for specialized depth cameras such as the Microsoft Kinect, while also increasing the distance from the camera in which the participants' gait can be analyzed.

Predicting parkinsonism severity in natural walking bouts on clinical scales remains a difficult computer vision and machine learning task. We note that there is a significant decrease in model performance compared to our previous work [22] in which models were fit and evaluated on all data rather than in a leave-one-subject-out cross-validation scheme. This confirms that there are large inter-person variations in gait, so models not trained on any data from a person they are being evaluated on will perform worse than if walks from that person are used in training. However, the LOSOCV scheme used in this manuscript is a more accurate measurement of model performance in a system that is pretrained and then deployed in a real-world setting.

Furthermore, discretizing gait impairment on coarse scales such as the UPDRS-gait and SAS-gait scales is challenging for both machine learning models and human annotators, particularly for walks which are on the border of two adjacent scores on the scales. Consistent application of the clinical scales by expert annotators is difficult in these cases, leading to inconsistencies in the ground truth scores [43]. This increases the difficulty of the prediction task, as the human annotator must use their best judgement with respect which score to assign, and the machine learning model must learn to predict scores in the same manner as the annotator's assigned labels. To minimize the impact of the variability in the ground truth labels, future work will focus on training models to predict clinical scores on continuous scales. Rather than evaluating these models with respect to how close the predicted scores are to the expert annotations, models will be evaluated by examining whether the relative scores of the walks are consistent with expert annotations (for example, walks with a manually annotated UPDRS-gait score of 0 should be predicted to have a lower score on the continuous scale than walks with a manually annotated UPDRS-gait score of 1). This work will focus on using expert labels as anchors for the clinical scales, while allowing models to predict scores with more granularity.

## V. ACKNOWLEDGMENT

The authors would like to thank Twinkle Arora, Melody Jizmejian, and Souraiya Kassam for their valuable help in data collection.